\documentclass[letterpaper]{article} 
\usepackage{aaai2026}  
\usepackage{times}  
\usepackage{helvet}  
\usepackage{courier}  
\usepackage[hyphens]{url}  
\usepackage{graphicx} 
\usepackage{natbib}  
\usepackage{caption} 
\usepackage{algorithm}
\usepackage{algorithmic}

\usepackage{newfloat}
\usepackage{listings}
\usepackage{times}
\usepackage{latexsym}
\usepackage{amsmath}
\usepackage{rotating}
\usepackage{subcaption}

\usepackage{multirow}
\usepackage{booktabs}
\usepackage{amsfonts}

\usepackage[hang,flushmargin]{footmisc} 
\usepackage{varioref}
\usepackage{cleveref}

\usepackage{xcolor}
\newcommand{\answerYes}[1]{\textcolor{blue}{#1}} 
 
\newcommand{\answerNA}[1]{\textcolor{gray}{#1}} 

\newcommand{\parai}[1]{{\it \noindent #1 }}
\newcommand{\parait}[1]{{\it \noindent #1 }}
\newcommand{\para}[1]{{\bf \noindent #1 }}
\newcommand{\eg}{e.g.,\ }

\title{Flattening Fantasies:
Analyzing Stereotypes in Pornographic Discourse\thanks{This paper contains analyses of pornographic language and discusses the perpetuation of ethnic, racial, and gender stereotypes and cultural taboos within that discourse. Some readers may find the language, themes, and discussions of stereotypes disturbing or offensive.}}
\author {
    Sarmad Chandio\equalcontrib,
    Osama Khalid\equalcontrib,
    Sharaf Zia,
    Ethan Kutlu 
}
\affiliations {
    University of Iowa\\
    \{sarmad-chandio, osama-khalid, sharaf-zia, ethan-kutlu\}@uiowa.edu
}

\begin{document}

\maketitle

\begin{abstract}

This study presents a large-scale quantitative analysis of 5.4 million video titles from xHamster, the world's fourth most visited pornography platform, to examine linguistic patterns used to describe different dimensions of identities (e.g., race and gender). Specifically, we investigate how nationalities, ethnic groups, and gender identities are portrayed in pornographic discourse, focusing on stereotypes and deviations from implicit norms of pornography. The results show the dominance of American pornography over other groups, which acts as the de facto norm for the adult industry. We identify persistent ethnic stereotypes reflecting colonial power dynamics, religious fetishization, and the sexualization of cultural taboos. The analysis reveals gendered representational differences across cultures, with a prevalent Madonna-Whore dichotomy in the portrayal of women and stereotypical depictions of men in physically dominant roles.

\end{abstract}


\section{Introduction}

Pornography and its consumption patterns have been subjects of academic research for over half a century \cite{laura1980take,marcus1966other}. Research on this topic has explored not only pornography's direct relationship to sexuality, but also its associations with violence, sexual assault, and relationship dynamics \cite{resch2014female}. With the proliferation of internet access, the use of pornography-related language is becoming more normalized and visible in mainstream discourse. Terms and concepts such as `MILF'(Mother I'd Like to Fuck), `thirst' (excessive sexual desire), and `NSFW' (Not Safe For Work), once confined to pornographic contexts, have gradually permeated the mainstream.

The digital era has ushered in unprecedented access to pornographic content, humorously encapsulated by the musical Avenue Q in the song ``The Internet is for Porn'' \cite{avenue_q}. This increased availability has sparked new discussions about the content and impact of pornography, bringing to the forefront concerns about representation and the perpetuation of harmful stereotypes. Psychological research has investigated these stereotypes through both laboratory and online studies, where participants report either their own experiences or their perceptions of others' beliefs regarding different forms of stereotyping \cite{greenwald1995implicit}. However, to date, no quantitative work has been done, at scale, on the language used in pornography to examine the nature of these stereotypes presented, and their prevalence across geographical and cultural spaces. This study seeks to address this gap. We argue that pornography, while ostensibly about deriving pleasure through observing sexual behavior, often provides a private space to explore taboos and fantasies where sexual behavior intersects with social representations of various ethnic, racial, and gender groups. In particular, it can reflect how marginalized and stereotyped groups including women, racialized communities, and sexual minorities, are depicted by dominant groups and portrayed in the content that is uploaded by content creators. As such, it provides a space for researchers to examine how individuals engage with stereotypes through language that is used to describe pornographic content. 

Earlier work has already identified how pornography can reduce women's roles to simplistic caricatures, erasing complexity and reinforcing damaging stereotypes, reducing them to one-dimensional objects of desire. This reductionist approach is considered a product of the `male gaze'. Film theorist Laura Mulvey introduces this concept to describe the depiction of women in visual arts and literature from a masculine, heterosexual perspective \shortcite{mulvey1975visual}. This framing is not limited to what men do but reflects a broader societal structure where patriarchal norms dictate how women are viewed and valued. Similar reductionist approaches are evident across history. For instance, the imperial gaze \cite{fischer2002imperial} privileged white women primarily for their reproductive roles while simultaneously hypersexualizing women of color, reducing diverse communities of color to essentialized, monolithic categories, erasing multilingual, multicultural identities and reinforcing otherness \citep{fanon2016black}. These dynamics have been widely documented across social science disciplines \citep{flores2015undoing}.

In pornography, this gaze manifests in the portrayal of women primarily as vessels for cis-men's pleasure, often devoid of agency or emotional depth \cite{dines2010pornland}. The male gaze in pornography not only objectifies women but also reinforces harmful gender stereotypes, presenting women as constantly available and willing participants in male sexual fantasies \cite{dworkin1981pornography}. 

This reduction can extend beyond sexual objectification to the imposition of rigid, binary tropes that reflect the Madonna--Whore Dichotomy, a psychological and cultural framework that categorizes women as either pure and submissive or sexually deviant. In pornographic discourse, this is mirrored through recurring titles and categories such as `teen', `innocent', or `schoolgirl', which emphasize youth, naivety, and chastity, contrasted with categories like `slut', `nasty', or `gangbang' which signal promiscuity and deviance. These roles carry distinct assumptions about women's sexual value and morality, and can reinforce harmful binaries that shape broader societal attitudes toward femininity and desire.


This flattening of gendered experiences also intersects with racial and ethnic categories \cite{cervulle2009queering, nagrath2003countering}. Rooted in colonial and post--colonial power dynamics, the imperial gaze reduces complex cultural experiences into simplistic, often fetishized categories like `Asian' or `Brown'. For instance, titles like ``\textit{amateur submissive Asian Wife [\dots]}''\footnote{Please note that all the titles used in examples are from our dataset.} portray Asian women as submissive yet hypersexualized, embodying a fantasy that combines docility with exotic allure. These are not isolated cases but part of a broader pattern repeatedly observed across our dataset.

In pornography, Black men are frequently portrayed through a racialized lens that emphasizes hypermasculinity and hypersexuality. These representations tend to exaggerate physical dominance, aggression, and sexual prowess. While these traits are superficially positive, however they reduce Black male identities to caricatures defined solely by physicality and virility. For example, titles like ``\textit{Big black man pounds [\dots]}'' exemplify this pattern, where Black men are repeatedly framed as overpowering, sexually aggressive figures. This persistent framing draws on long-standing racial stereotypes and contributes to a dehumanizing narrative that equates Black male worth with sexual performance and physical domination \cite{williams2004skin}.

Such stereotypes, while considered racially insensitive in mainstream discourse, are often amplified in pornography, and can potentially translate into real-world violence and hate crimes \cite{reed_21,campbell2023influence}. Prior research has shown that the language and tropes found in pornography closely mirror those used in misogynistic online communities, such as involuntary celibate or `incel' groups \cite{tranchese2021don} . 

The potential for pornographic content to shape societal attitudes and reinforce harmful stereotypes should be understood within the broader media ecosystem, underscoring the importance of systematically studying how different racial and gender categories are represented within pornographic content. Given its widespread and often private consumption, the question becomes: what is the incidence of pornographic stereotypes (i.e. can text data substantiate the presence of stereotypes posited in research noted above), and what is their prevalence (i.e. how widespread is the usage of stereotypical language and terms with demographic and ethnic groups)?

While pornography is primarily consumed as visual media, the experience is mediated by textual and linguistic elements such as search terms, tags, descriptions, titles, etc. These components serve as entry points and framing devices for consumers and, as prior research suggests, may contribute to the perpetuation of stereotypes and the simplification of complex sexual dynamics and identities \cite{ramirez2022social}. Rather than treating individual titles as isolated acts of representation, our approach identifies recurring linguistic patterns that position social groups, contributing to broader discourses that normalize and reinforce stereotypical portrayals. Importantly, this research can provide a venue for scholars to investigate what type of processes can be implemented to reduce the perpetuation of these stereotypes and biases in online spaces.

We investigate how gendered and ethnic stereotypes manifest in text by analyzing pornographic video titles. Specifically, we ask the following questions:

\begin{itemize}
\item \textbf{RQ 1:} 
What emergent clusters of ethnic representation arise from patterns in the language used in pornographic video titles?

We propose a title-level representation approach to quantitatively analyzing pornographic content. This method allows for a systematic comparison of linguistic patterns in adult content grouped by video-level nationality tags that include labels, such as `India', `German', or `American' that are used by the platform to organize content, into clusters that broadly coalesce across ethnic categories.

\item \textbf{RQ 2:} How are ethnic stereotypes and cultural taboos represented in language?

We investigate the ways in which linguistic choices in pornographic video titles reflect and engage with ethnic stereotypes and cultural taboos. Our analysis highlights recurring patterns of ethnic exoticization, where identities are represented through simplified, sexualized, and often racialized language. These representations vary across cultural contexts, revealing how particular words, phrases, and themes are used to signify otherness, taboo, or fantasy. This approach allows us to examine how stereotypes are encoded and circulated through the language of pornographic metadata, without presupposing intent or effect.

\item \textbf{RQ 3:}  How are gender roles represented in pornographic titles across different cultural contexts?

We examine how gendered language is used in pornographic titles across different cultural contexts, focusing on the terms and descriptors associated with masculinity and femininity. By analyzing the semantic relationships between gendered terms within each nationality group, we observe cross-cultural patterns and divergences in how male and female gender roles are constructed. This includes variations in sexual agency, physical descriptors, and relational framing, offering insight into how gendered identities are linguistically encoded within pornographic discourse.
\end{itemize}

\vspace{-0.7em}
\section{Related Work}
The study of pornography is an active, multidisciplinary field of research \cite{haywood2024becoming,nocella2024stigma,franco2024fucking,cole2014pornography,virdis2015eroticizing,moorman2024bi,thorneycroft2023choosing}. Much of the research on pornography within Computer Science has focused on network analysis and content distribution. Works such as those by  \citet{morichetta2021understanding} and \citet{ahmed2016internet} examine HTTP logs to explore patterns in porn traffic across networks.
Unlike these approaches, our work aims to examine linguistic patterns in published content, specifically how language is used within pornography to socially construct identity categories and associated stereotypes.  

\subsection{Pornography and Language}

Mulvey's foundational work \shortcite{mulvey1975visual} on the `male gaze' in cinema has been instrumental in explaining how visual media, including pornography, often reduces women to objects of male pleasure. Building on this concept, Pihlaja's study \shortcite{pihlaja2016expressing} analyzed around 22,000 comments from the 100 most popular videos on Pornhub. This research revealed that while comments were generally not overtly offensive, they often contained negative descriptions of women, reflecting the objectifying nature of the male gaze in pornographic discourse.

Mach's work \shortcite{mach2022smitten}, similar to ours, explored the language of pornography by analyzing a corpus of 17,000 Pornhub video titles, focusing on gender representation and the roles assigned to men and women. Specifically, they examined how agency is constructed through verb-object relationships (primarily using the verb `fuck') and how roles are linguistically encoded using nouns and adjectives associated with each gender. Their method relied on part-of-speech tagging. While useful for quantifying grammatical features (e.g nouns, verbs etc.), this approach has limited capacity for extracting semantic meaning. More germane to our work, is prior research that has looked at emergent patterns of language and meaning from text on the internet. Caliskan and colleagues \shortcite{caliskan2017semantics} found through a large-scale language corpora that the internet holds the imprints of our historic biases such as associations between gender and professional roles. Crucially, they showed that these biases are not inert; they are replicated and even amplified in algorithmic systems. Subsequent research has shown that algorithmic systems such as large language models (e.g., ChatGPT) can exhibit gender bias in tasks like machine translation, particularly in languages with gender-neutral pronouns \cite{ghosh2023chatgpt}. These findings highlight the broader concern that language used in digital environments does not merely reflect social stereotypes but can also amplify them at scale \cite{noble2018algorithms,habib2024algorithmic}.

\subsection{Pornography and Tags}

While our study focuses on the language of video titles, we use tags, specifically nationality tags, as a way to organize and group the content for analysis. Tags function as metadata that assign categorical identity to videos (e.g., `Indian', `German'), enabling us to compare linguistic patterns across these tagged groups.

Wong et al.'s study \shortcite{wong2020youtube} that analyzed video tags for 3 million xHamster videos. By representing tags as a network of relationships, they observed a hierarchy in which some tags are more frequent and dominant while others are niche. However, this study did not examine how tags relate to questions of representation, gender, or race.
Similarly, others quantitatively analyzed tags on XNXX and xHamster to explore how nationalities were represented through co-occurring tags \cite{mazieres2014deep}. We extend this work by using nationality tags as an organizational framework to examine the linguistic construction of stereotypes in video titles.

 \section{The Pornographic Ecosystem.}
\subsection{How Pornographic Content Is Produced, Distributed, and Categorized Online}
The pornography industry is a multifaceted and complex ecosystem with various players, each contributing to the production, distribution, and consumption of adult content.

 \parait{Producers} 
 in the porn industry include both individual content creators and large production studios like Faphouse and BangBros. These entities have the most autonomy during production, deciding on the contents' themes, performers, and overall direction. 

 \parait{Consumers} are the people who view pornography. Their preferences and viewing habits significantly influence the types of content produced and made available on various platforms.

\parait{Platforms}
such as Pornhub, xHamster, and XNXX dominate the distribution of pornographic content. These websites serve as the primary venues where consumers access and view videos.

\parait{Uploaders}
 are a significant part of the ecosystem, with most videos on these platforms being uploaded by pseudonymous users rather than the platforms or producers. These uploaders often re-upload existing videos with new titles, tags, and categories, reflecting their own interpretations and market demands.

\para{The anatomy of a video:} In addition to video content, a pornographic video includes associated metadata such as the date of upload, view counts, duration, and comments, along with its title, tags, and categories. The title is a free-text field that reflects the uploader's description of the video. Tags are specific keywords or phrases that describe details or aspects of the content, such as themes like `threesomes', performers, or scenario-based content like `plumber'. Categories are broader classifications that group content into general themes or genres, such as `Amateur', `Professional', or `Popular' as well as nationalities `Chinese', `Danish' and racial identities `Latino'. These fields help organize the content library on the websites, making it easier for users to navigate and find content related to their specific interests.

\section{Dataset Description} \label{sec:dataset}

\begin{figure*}
    \centering
    \includegraphics[width=1\textwidth]{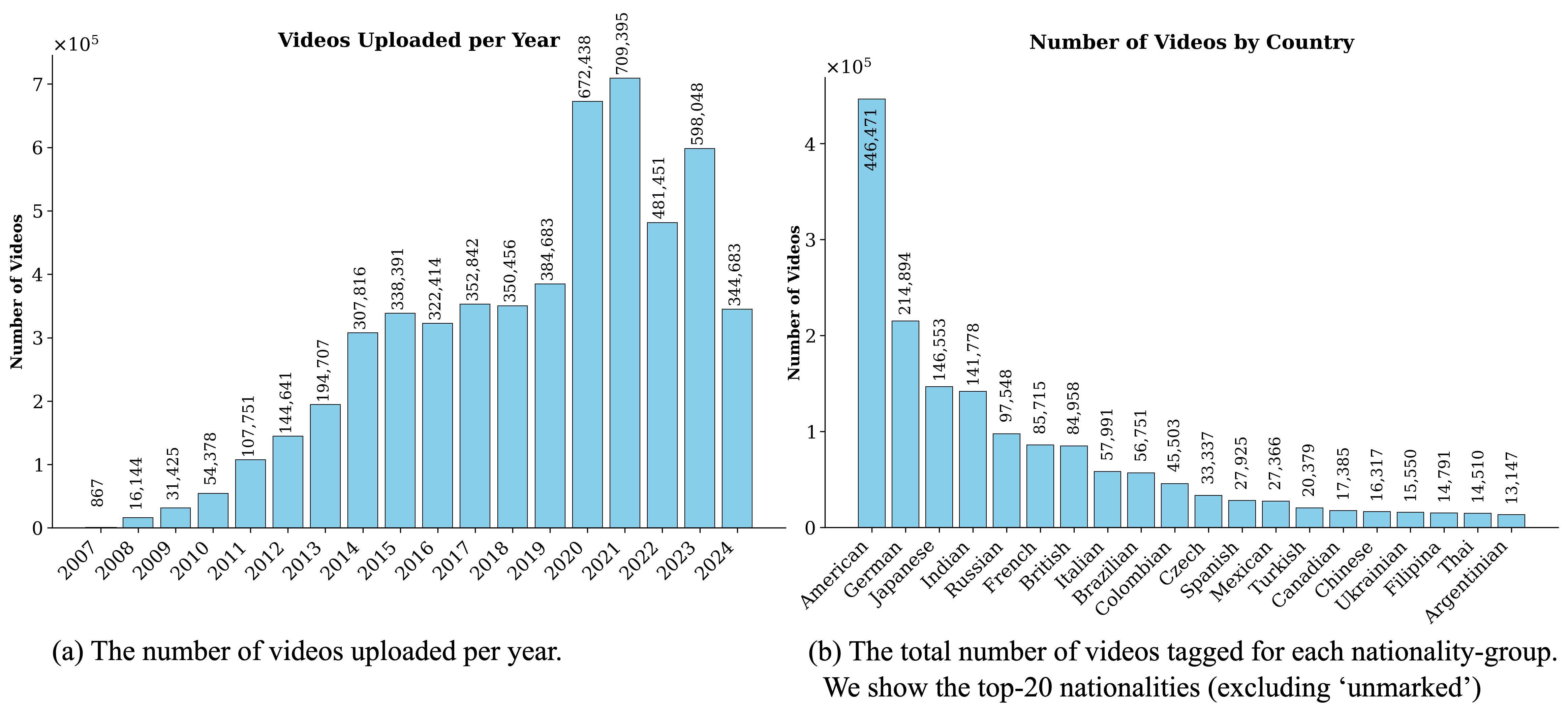}
    \caption{Breakdown of the trends observed in the dataset.}
    \label{fig:three_subfigs}
\end{figure*}

We utilize publicly available data from xHamster, the fourth most visited pornographic website globally in 2024 \cite{semrush2024trendingwebsites}. Our dataset, collected in July 2024, comprises metadata for 5,412,529 videos, including titles, tags, descriptions, upload dates, engagement metrics, and duration.
\Cref{fig:three_subfigs}(a) summarizes the rate of video uploads over time. The data shows a steady, linear upload growth until 2019, followed by a sharp spike in 2020 and 2021, likely driven by increased online activity during the COVID-19 pandemic \cite{zattoni2021impact}. This surge peaks in 2021 with 709,395 uploads before dropping in 2022 and 2023 as user behavior returns to pre-pandemic trends. The 2024 data, representing uploads for six months (from January to July).

Out of the 5.4 million videos, 3,547,673 were not tagged for any nationality\footnote{Nationalities are explicitly tagged in the xHamster data structure.}, and 1,841,860 had exactly one nationality tag. The remaining videos had two or more nationalities tagged. 
The most common nationality was tagged American with 450,865 videos, followed by German and Japanese (215,989 and 146,914 videos, respectively) shown in \Cref{fig:three_subfigs}(b). We note nationalities follow a long-tail distribution. In our analysis, we focused on nationalities that had at least 1000 videos. The resulting set of 71 nationalities\footnote{In our analysis, we treat the Unmarked videos as a separate nationality} represents 99.5\% of the tagged videos.

\begin{figure}[h!]
    \centering
    \includegraphics[width=0.47\textwidth]{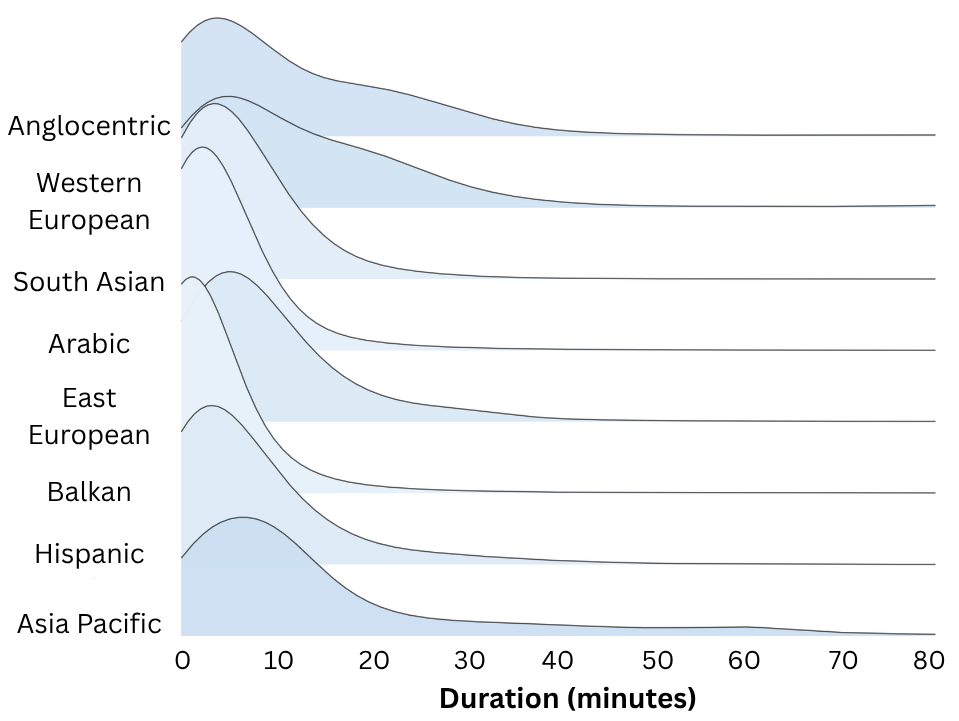} 
    \caption{The distribution of video duration for each ethnic cluster. All clusters had a mean video length $<$ 20 minutes. }
    \label{fig:sub:ridge_plot_time}
   
\end{figure}

\Cref{fig:sub:ridge_plot_time} shows the upload duration of videos from various groups (discussed in section \Cref{sec:dataset}). Across all groups, we observed that video durations were predominantly within the 5-20 minute range.

 This observation is consistent with data from other major pornographic platforms; for instance, PornHub reports an average time spent on the website is 9 minutes and 20 seconds \cite{pornhub2024longshort}. This finding is also consistent with previous works that found the average session to be 15 minutes or fewer \cite{morichetta2019characterizing}.
 
 While all groups show some degree of right-skewed distribution (indicating the presence of longer videos), the extent of this long tail varies notably between groups.
The most prominent long tail is observed in the Asia Pacific group. An extended tail is also visible in the Western European group. This extended right tail indicates a higher proportion of lengthier videos than other groups. The remaining groups show a relatively compact distribution adhering to shorter, more standardized videos.

\section{Empirical Findings} \label{sec:methods}
\subsection{Representing Nationalities and Titles}
We define videos based on the nationalities they are tagged for by the uploaders. We group videos into nationality-groups based on these tags. For our analysis, we study the language of video titles corresponding to different nationality-groups.

Each video in the pornographic dataset has a title and one or more tagged category\footnote{The tags/categories are predefined by xHamster and uploaders select from these options to categorize the uploaded video.}. Of the 981 unique categories, 182 specifically correspond to nationalities, which we use to define our nationality-groups. A video tagged for nationality makes the performer’s perceived national identity a central focus, explicitly associating the content with a specific ethnic or cultural background.

Our analysis represents each nationality-specific group using the language in video titles. The free-text of user-generated titles often contains nuanced or descriptive language offering insights into how users and content creators frame their content. We quantify the association between a group $c$ and a word $w$ in video titles by representing all video titles that belong to a group as a single pseudo-document and calculating the TF-IDF \cite{robertson2004understanding}:
\begin{equation*}
\text{TF-IDF}(c,w) = \text{TF}(w|c)\cdot \text{IDF}(w)
\end{equation*}
Where $\text{TF}(w|c)$ is the frequency of the word $w$ in all the videos of $c$ and $\text{IDF}(w)$ is the inverse-document frequency of $w$ in all the groups. 
We represent each group $c$ as a vector of TF-IDF values of the group and all the words in the vocabulary of all the preprocessed titles\footnote{The titles were preprocessed using standard NLP operations i.e. lowercasing, removing English stopwords, and lemmatization.}.

To illustrate how this method captures group-specific linguistic patterns, consider the following example: if the word `maid' appears frequently in titles tagged with `French' but rarely in titles from other nationality groups, it will have a high TF-IDF score for the `French' group. This would indicate that `maid' is particularly characteristic of how `french' content is linguistically framed. In contrast, a word like `sex' may appear frequently across all groups and thus have a high term frequency but a low TF-IDF score. As such it lacks discriminatory power and does not help distinguish the linguistic framing of any particular group. TF-IDF helps isolate language that is specific and meaningful to a group's representation, rather than just commonly used across all content.

Importantly, this representation accounts for the varying number of videos within each group by inherently adjusting for document frequency across the entire corpus. 

Not every video is tagged with a nationality, and the absence of such tags suggests that the performer’s national origin is not emphasized or is treated as default and unmarked. In our analysis, we consider the videos tagged for nationalities and those that are not tagged, allowing us to explore how the presence and absence of explicit nationality may influence representations in the dataset. To minimize tag-spamming\footnote{tag-spamming: excessive or irrelevant tagging.}, we limit our dataset to only include videos with either one or no nationality tags.

\subsection{Clustering of Nationalities}

We represent a nationality-group as a vector of the TF-IDF values of all the words across all nationalities. We use Ward’s minimum variance method to cluster \cite{ward1963hierarchical} the TF-IDF vectors. 

Figure \ref{fig:combined-analysis} shows the hierarchical clusters of the TF-IDF vectors. These clusters indicate emergent ethnic categories of shared cultures and geography. Expectedly, nationalities in geographical proximity in the real world (\eg Pakistani-Indian and German-Swiss) are clustered closer to each other based on their title descriptions. In some cases, however, nationalities cluster not due to geographical proximity but due to similarities in the way they are described in pornographic discourse. For example, American and Australian titles cluster closely, suggesting a convergence in the sexualized language used to represent these groups. We refer to such groupings as ``culturally similar'' in the sense that the titles evoke overlapping stereotypes, vernaculars, or themes, as identified through unsupervised clustering.

We can broadly categorize the clusters into 8 groups based on either language or region:
Anglosphere (American, Australian, Canadian, Nigerian), Western European (German, Swiss, French), South Asian (Indian, Bangladeshi, Pakistani), Middle-East and North Africa --- MENA (Moroccan, Saudi, Egyptian), Eastern European (Russian, Ukrainian ), Balkan (Serbian, Croatian, Turkish), Hispanic (Spanish, Mexican, Argentine), and Asia Pacific (Japanese, Chinese, Malaysian, Indonesian).
To explore how language is used to describe ethnic categories in pornographic content, we focus on five nationalities as exemplars: French (Western European), Chinese (Asia Pacific), Saudi (MENA), Mexican (Hispanic), and Indian (South Asian). These exemplars were selected to represent a wide range of cultural, geographic, and linguistic clusters, while also reflecting nationalities with substantial representation in the dataset and distinctive stereotype patterns. Together, they allow for a comparative analysis across dominant cultural tropes in different regions.

\begin{figure*}[ht]
   \centering
   \includegraphics[width=\columnwidth, angle=90]{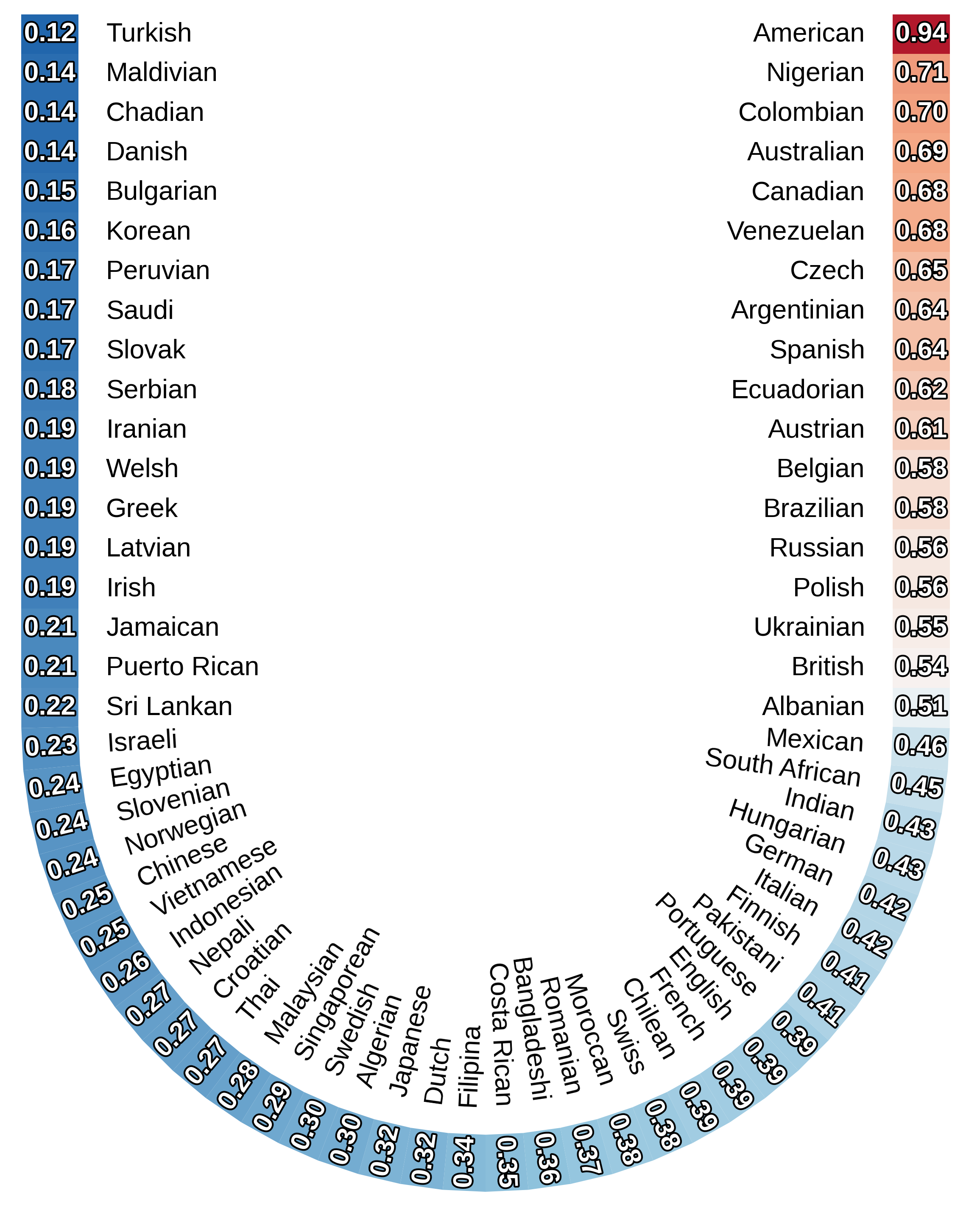}
   \caption{Heatmap of similarities of the title-level vector representations with the unmarked default.}
   \label{fig:dendo-titles}
\end{figure*}

\subsection{Cultural and Gendered Differences in Porn}\label{sec:gender_diff}

\para{Cultural differences.} 
Cultures and cultural identities are complex constructs that can not be modeled simplistically. We use nationality tags as proxies of cultural representations. Even though this model is not a perfect reflection of culture, it provides us with a useful lens to study representations in pornography.

We investigate whether the representation of a particular nationality-group $c$ significantly differs from the general pornographic representation. Our null hypothesis posits no difference between the representation of a specific group and the overall space of pornographic videos. In that case, we would expect the distribution of the group's representation to be indistinguishable from the general universe of pornographic content.

\parait{Method.} We represent the general pornographic universe by constructing a `random video title-set' that samples videos from the set of all pornographic videos. We construct a random video title-set as follows: For each group $c$, we consider a set of $n$ associated videos $V^c = {v^c_1, v^c_2, \ldots, v^c_n}$. We then construct a corresponding set of random videos $\hat{V}^c = {\hat{v}^c_1, \hat{v}^c_2, \ldots, \hat{v}^c_n}$, where each $\hat{v}^c_i$ is sampled from the general distribution of videos in the pornographic universe. The general distribution is derived from the set of all videos across all countries, defined as $\hat{V} = \{ V^c \mid c \in C \}$ where $C$ is the set of all nationality-tagged videos in our dataset. 


\parait{Evaluation.}
We construct random representations of titles for each group using the above process. For any group $c$ with a vector representation $r^c$, we generate $m$ random representations $\{\hat{r}^c_1, \hat{r}^c_2 \cdots \hat{r}^c_m\}$. For each vector $r_i \in \{r^c\}\cup \{\hat{r}^c_1, \hat{r}^c_2 \cdots \hat{r}^c_m\} $, we calculate its average cosine similarity with all other elements. We then rank these vectors in decreasing order of average similarity.

To determine statistical significance, we check whether the real vector $r^c$ has a lower average similarity than at least 95\% of the vectors in $\{r^c\}\cup \{\hat{r}^c_1, \hat{r}^c_2 \cdots \hat{r}^c_m\}$ (i.e. p-value < 0.05 ). This approach allows us to assess whether a group's representation significantly differs from random expectations within the pornographic video space. 

\para{Universal divergence from random.}
We compare the language of the titles of nationality-tagged videos (including Unmarked) with a random group created following the approach outlined in \Cref{sec:gender_diff}. All the videos in each random group are considered part of the same pseudo-document. We use these pseudo-documents to create TF-IDF vectors for the random groups.

We find that the titles of all nationality-tagged groups are significantly different from random groups (p-value $<$ 0.05). This universal divergence suggests that pornographic content, regardless of nationality, follows specific patterns or conventions that distinguish it from random text. 
This finding underscores the existence of a global `language of pornography'.

While all nationalities show significant differences from random groups, the magnitude of these differences varied considerably. This variation is visualized in \Cref{fig:dendo-titles}, which shows the cosine similarities of the TF-IDF vectors of nationality-tagged groups with random titles. The varying magnitudes of similarity with random groups can be interpreted as a measure of how closely each nationality's pornographic discourse aligns with a global or mainstream pornographic language. Nationalities with lower similarity to random groups may be using more culturally specific language or themes, while those with higher similarity may be more aligned with global pornographic conventions.

\para{The centrality of America in pornography.}

Out of the four most visited pornographic websites (YouPorn, PornHub, RedTube and xHamster), three are owned by the same North American company, Aylo \cite{aylo2024brands}. The sole exception is xHamster, which is owned by a Cypriot company founded in Czechia and reportedly controlled by Russian nationals \cite{jacquemien2021operators}. Even though it is a European company, the videos on xHamster are predominantly tagged American (\Cref{fig:three_subfigs}b).

\begin{figure}[p] 
    \centering
    \begin{minipage}{0.80\columnwidth}
        \centering
        \includegraphics[angle=-90, totalheight=0.90\textheight, width=\linewidth, keepaspectratio]{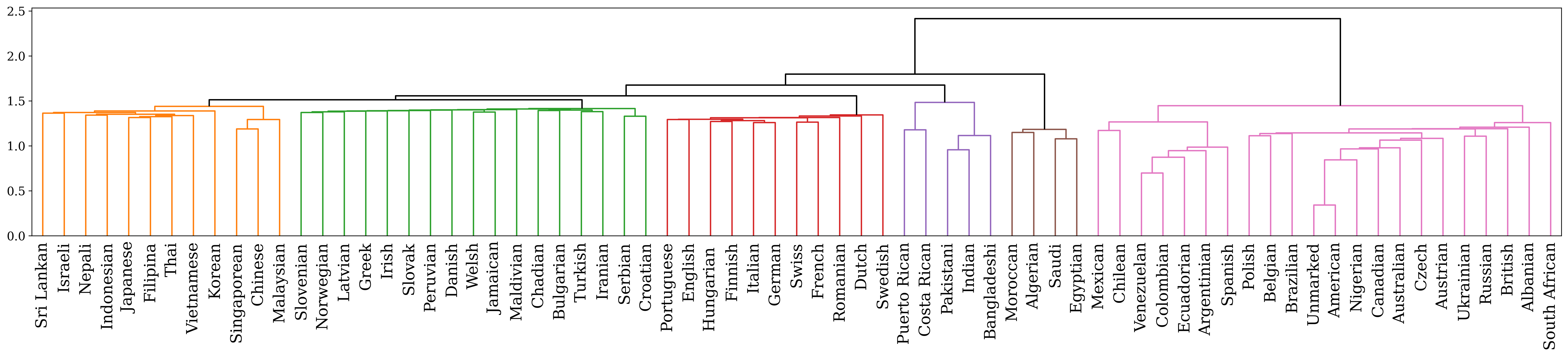}\\[0.2cm] 
        (a)
    \end{minipage}\hfill 
    
    \caption{Dendogram of title-level vector similarity. Lower branch connections indicate greater similarity. Cutting the tree at different levels reveals various clustering arrangements}
    \label{fig:combined-analysis}
\end{figure}

The `American' group showed the highest similarity (following `Unmarked' at 0.98) to the random category (cosine similarity = 0.93), indicating the centered-ness of the category in pornographic discourse (\Cref{fig:combined-analysis}). The high similarity of `American' can not be attributed to its volume alone. We find `German' --- the $3^\text{rd}$ largest group by volume (4\% of total videos) --- only had a 0.53 cosine similarity with the random universe, whereas `Canadian' the $16^{\text{th}}$ most populous (0.3\% of total videos) category had a much higher similarity (0.80). 

\para{The globalization of pornographic language.}
The general pattern reflects a gradient of `Westernization' in pornographic discourse, with `Unmarked' and `American' as the defacto norms. This pattern, however, is not uniform across all nationality-tagged groups. 
Some non-Western nationalities showed high similarity to the random category, notably `Colombian'(0.72) and `Nigerian' (0.71), suggesting the adoption of Western pornographic language norms. For these two groups, among the top-20 words with the highest TF-IDF scores in these groups, most were sexually explicit generic terms like \textit{`hot'}, \textit{`sex'}, and `\textit{fuck'}, with only \textit{`nigeria'} and \textit{`latino'} hinting at local content.


On the other hand, Asian and Middle-Eastern nationalities predominantly use localized language in titles. This reinterpretion and reimagining of pornographic discourse into local vernaculars manifest in the use of region-specific terms, \textit{`fuck'} is replaced by `\textit{chadai}' in the `Indian' context, \textit{`sex'} with `\textit{entot}' in the `Indonesian', \textit{`blowjob'} with `\textit{sakso}' in `Turkish' discourses.

We hypothesize that deviations from the (Western) norm can primarily be attributed to two factors,  First, using local languages in titles, indicating that video uploaders are likely native speakers or fluent in these languages. Second, the development of culture-specific pornographic genres and terminology. One example is the relatively high use of `\textit{hentai}' with `Japanese'.





\subsection{Ethnic Stereotypes}

Ethnic stereotyping in pornographic titles can be divided into two categories: outgroup stereotyping and ingroup stereotyping. Outgroup stereotyping manifests as `Otherization' and involves ethnic fetishization, while ingroup stereotypes engage with culturally specific taboos and their subversion.

\para{Outgroup}
\\
\parait{Colonial Otherization.}
Orientalism, according to scholars \cite{said1979orientalism}, is fundamentally about the power to define and represent outgroups, racial and ethnic categories distinct from one's own. 
The language of pornography also represents an orientalist gaze that reinforces the relationship between the colonizer `Self' and the colonized `Other'. This is apparent in the language used in the pornography of the former colonizer (`French') and the formerly colonized (`Arab'). A prominent example is the use of `beurette' in both the Saudi and French sets. 

This French slang describes female immigrants from the Arab Maghreb. It appears in video titles like:
\textit{``danseuse marocaine arab beurette fesse ass'' (trans: Moroccan dancer arab beurette spanks ass)} 

%
Some scholars argue that Maghrebi women have been fetishized since the start of the French colonial rule in North Africa \cite{hafsa}. French colonizers in Algeria categorized prostitutes by their ethnic and physical characteristics. Such characterization continues to this day and is represented in and reinforced by the language of pornography we look at for French porn. Such stereotypical representation continues to dehumanize women from minority groups to satisfy a dominant ingroup's sexual fetishization of a minority outgroup \cite{taraud2011prostitution}. 


\parai{Religion.} Religious symbols and identities are similarly appropriated and sexualized in porn, particularly within Middle Eastern contexts. In the `Saudi' set, terms like `\textit{hijab}' and `\textit{muslim}' were in the top-20 words when ranked by TF-IDF scores. They were used in titles like \textit{``hijab arab lady [$\cdots$]''} and \textit{``shameless muslim hijabi\footnote{`Hijab' is Arabic for veil. `Hijabi' is a person who wears hijab.} [$\cdots$]''} suggesting a fetishization of religious taboos and the intersection of cultural and religious identity within sexual content. The rise of 'porno ethnik' (ethnic porn) in cinema, and the use of hijab as a form of striptease, can be thought of as a by-product of colonial repression and ridicule of symbols deemed incompatible with Western ideas of progress \cite{mack2017sexagon}.

Such religious fetishization indiscriminately orientalizes and codes performers from the Middle-East as Muslim. For example, titles like \textit{``muslim pornstar mia khalifa sucks [$\cdots$]''} code Mia Khalifa as Muslim, based on her brownness, disregarding her actual Christian identity for the sake of the video \cite{bella2018mia}.

The ethnic exoticization observed, especially in the context of former colonial relationships, highlights how historical power dynamics continue to shape sexual representation and fantasy. The fetishization of cultural and religious symbols, such as the hijab, further underscores the complex interplay between sexuality, culture, and power. The focus on the hijab not only exoticizes the Other but also mirrors the objectification found in the sexualized exposure of women’s nudity. While one renders power through concealment and symbolic cultural markers, the other does so through exposure and bodily availability. Both function as mechanisms of objectification that reduce women's bodies to signifiers of erotic difference. These findings raise important questions about the role of pornography in perpetuating harmful stereotypes and colonial tropes. They also highlight the need for more critical engagement with pornographic content, as a reflection of broader societal attitudes towards gender, race, and sexuality. 

\begin{figure*}[htb]
   \centering
   \includegraphics[width=1\textwidth]{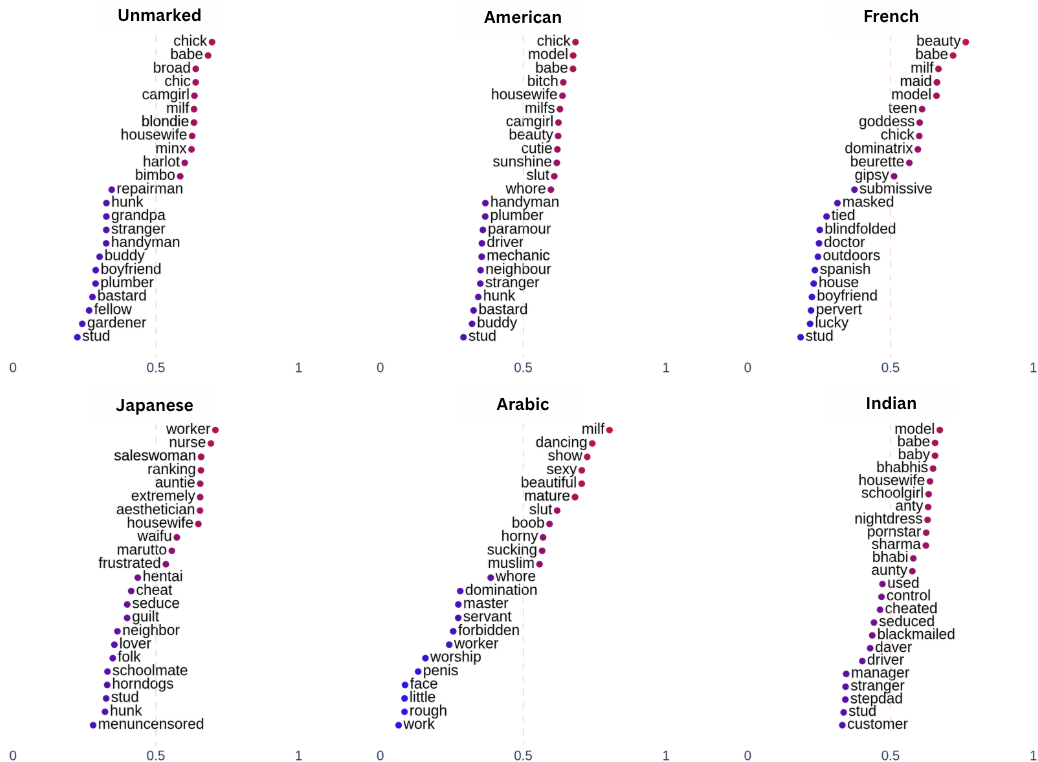}
      \caption{The gender association ratio $G(w)$ of a sample of the words (discussed in \Cref{sec:gender_diff}) of the 6 exemplars (we use Arabic nationality-group due to the small size of its exemplar). Words above 0.5 share a greater context to the feminine vector, and words below 0.5 share a greater context to the masculine vector. For instance, in the Unmarked category (top-left), the word \textit{`stud'} ($G(w)=0.28$) shares a greater context with the masculine vector than the feminine vector and vice versa for the word \textit{`chick'} ($G(w)=0.70$).}
   \label{fig:log-odds}
\end{figure*}

\para{Ingroup}
\\
\parai{Cultural specific taboos.} 
 Previous studies have argued that pornography is a domain where otherwise taboo sexual fantasies are realized, and it is used as a tool to release pent-up sexual repression \cite{ellis2006pornography, mondin2017tumblr}.
We find sexualization within kinship relations and domestic settings as a manifestation of taboo fantasies. As an example, within the `Indian' category, terms like `\textit{bhabi}' (sister-in-law) or `\textit{kaamwali}' (maid) frequently appear in titles, as in \textit{``bhabi cheating [$\cdots$]''} and \textit{``Indian Kaamwali Maid Fucked by [$\cdots$]''}. This sexualization of familial and household roles reflects a distinct form of exoticization that plays on cultural taboos.

However, this exoticization is not only specific to the Indian context. We also find this phenomenon of sexualizing visible in other cultures. Pornographic categories tagged `\textit{mom}' in `American' and `\textit{maid}' in `Japanese' or `French' cultures (\Cref{fig:log-odds}) represent similar taboo fantasies and serve similar purposes, invoking an immediate caretaker fantasy for men fantasizing about sex with women who are in close physical space to them.

We find the pervasiveness of patterns of ethnic stereotyping and exoticization within pornographic discourse, as evidenced by the use of racialized language in the set of the top-20 TF-IDF words for nationalities like `French' and `Saudi'. This exoticization is not only limited to outgrouping or racialization, it also presents itself as an artifact of cultural repression and fantasies as illustrated by kinship and caretaking themes in various communities, including but not limited `Indian', `Japanese', and `American'.

While the findings above do not fully address the extent to which pornography reinforces and potentially amplifies societal biases related to race, ethnicity, and culture, the observed incidence and prevalence of stereotype-rich language underscores the need to critically examine how ethnic and cultural identities are represented in adult content.

\para{Gendered differences.}
Pornography, as a reflection of societal attitudes and desires, often presents exaggerated and stereotypical representations of gender. While sexualization is prevalent across all genders in pornographic content, the nature and extent of this sexualization can vary significantly between masculine and feminine portrayals. We explore and quantify these differences, shedding light on the ways in which gender is constructed and presented in adult content across different cultural contexts. Due to the limitations of the data collection, we used man/woman dichotomy to represent gender differences and, therefore, were unable to implement non-binary identities.

\parait{Method.} We use Word2Vec \cite{mikolov2013distributed} to create word-embeddings of the pornographic video titles from each nationality-group in our dataset. These embeddings capture the semantic relationships between words and concepts specific to each cultural context. Titles serve as concise summaries of content, capturing key themes, performers, and actions in a highly condensed format.

These embeddings allow us to quantify semantic relationships by measuring vector similarities between pornographic concepts and gendered reference vectors. For instance, if we find that the word ‘maid’ is in closer proximity to the feminine vector compared to the masculine vector, this would indicate that the term is more semantically aligned with femininity in the context of pornographic discourse.

We use previous methods from studies to create representative gender vectors by averaging the embedding vectors of gender-specific words and pronouns\cite{bolukbasi2016man}. For the masculine dimension, we average the embeddings of 
the masculine words, while for the feminine dimension, we use the embeddings of the feminine words\footnote{masc. words = ['man', 'he', 'his', 'him', 'lad', 'boy', 'men', 'guy', 'male'] \\
fem. words = ['woman', 'she', 'her', 'female', 'lady', 'girl', 'lass']}. 
Given a word $w$ in our embedding-space, we represent the relative gender association $G(w)$ as : 
\[G(w) = \frac{\cos(\vec{w}, \vec{m})}{\cos(\vec{w}, \vec{f}) + \cos(\vec{w}, \vec{m})}\]

\noindent where $\cos(\vec{w}, \vec{m})$ and $\cos(\vec{w}, \vec{f})$ represent the cosine distances of the embedding of $w$ from the average masculine and feminine vectors, respectively. 

\parait{Evaluation.} The $G(w)$ ratio ranges from $0$ to $1$, where values closer to $1$ indicate stronger feminine associations, values closer to $0$ suggest stronger masculine associations, and $0.5$ represents equal association with both genders. 

\para{Unmarked and American portrayals}
In the `Unmarked' category, feminine descriptors predominantly emphasize physical attributes. Conversely, masculine descriptors often relate to occupations traditionally associated with manual labor or technical skills (\eg \textit{``Blonde MILF wants a plumber to fuck her''}). The unmarked baseline establishes a general trend of objectification for women and occupational identity for men in pornographic content. 

The representation of men in `American' follows the same pattern as the `Unmarked' category and is closely associated with labor-intensive professions. In contrast, women follow an appearance-heavy slant where words such as  \textit{`beauty'}, \textit{`blondy'}, \textit{`model'}, and \textit{`chich'} generally refer to their appearance or hypersexualized character and are used in titles like \textit{``Beautiful slattern gives ... her [$\cdots$]}''.

\para{Western European depictions.}
Within the Western European group, we find associations between French women and the word `\textit{maid}'. While the fetishization of the maid reflects a power dynamic emphasizing a master-servant relationship \cite{apter20188}, in the `French' context, as noted earlier, it also demonstrates a racialization of the power dynamic as the `maid' trope is being used in conjunction with racialized groups ("\textit{French Arab Muslim maid}").

On the other hand men are linked with terms such as `\textit{pervert}', `\textit{tied}', `\textit{masked}', and `\textit{submissive}', that suggest a prevalence of BDSM themes and a subversion of traditional masculine dominance in French pornography, as hinted by titles like (\textit{``french dominatrix fuck submissive husband [$\cdots$]''; ``tied hand stepbrother [$\cdots$]'')}

\para{Asian portrayals.}
In the `Japanese' set, we observe a distinct characterization of male roles that focuses on themes of moral transgression and power imbalances, and is indicated by association of words such as \textit{`cheat'}, \textit{`seduce'}, \textit{`neighbor'}, and \textit{`guilt'} leaning towards the masculine ($G(w)<0.5$).

These themes manifest in titles such as \textit{``[$\cdots$] cheating husband''} and \textit{``[$\cdots$] teacher seduces schoolgirl''}, highlighting a prevalent narrative of men in positions of trust or authority violating social norms or ethical boundaries. The \textit{``teacher seduces schoolgirl''} trope, in particular, underscores a problematic power dynamic where men in authoritative roles exploit their position to initiate sexual encounters with women subordinates or dependents. 
This representation contrasts with, and complements, the portrayal of women in the Japanese set, who are often depicted in service-oriented or submissive roles like `nurse' and `saleswoman' in titles like 
(\textit{``Door-to-door saleswoman offers more than just goods''; ``beautiful japanese nurse [$\cdots$]''}).

\para{Madonna-Whore Dichotomy.} 
The Madonna-Whore Dichotomy (MWD) is a psychological framework that categorizes women into two mutually exclusive groups: the virtuous, pure `Madonnas' and their sexually promiscuous counterparts, `whores'. This dichotomy reflects a deeply problematic view of female sexuality and personhood \cite{bareket2018madonna}. Finding evidence of MWD in pornographic content is significant because it suggests that these materials can not only reflect but potentially reinforce harmful stereotypes about women. This can contribute to the objectification of women, difficulties in forming healthy relationships, and the perpetuation of sexist attitudes in society at large \cite{kahalon2019madonna}. The presence of MWD in pornography across different cultures indicates that this is a widespread issue and not a localized issue for particular sub-groups. 

Universally, we find sexist language targeted against women across almost every group. This includes derogatory terms like `sharmota' (Arabic for whore) in the Saudi set and `prostituta' (Spanish for prostitute) in the Mexican set. We also observe sex acts indicating dominance over women, such as `facialisee' (French for the sex-act: facialized) in the French set.

Interestingly, kinship terms are also sexualized, like `Mertua' (Indonesian for Mother-in-Law) or `Bhabi' (Hindi for sister-in-law). This dichotomy of women being sexualized and dehumanized as `bad', `promiscuous', and `seductive whores', as well as the use of kinship terms to indicate them as either `pure' and `chaste', signals the presence of a Madonna-Whore Dichotomy in pornographic language \cite{bareket2018madonna}.

\section{Discussion}
In this study, we present a quantitative analysis of 5.4 million video titles from xHamster to examine linguistic patterns used to describe different dimensions of identities (e.g., race and gender). Specifically, we investigate how nationalities, ethnic groups, and gender identities are portrayed in pornographic discourse, focusing on stereotypes, their nature and prevalence across pornographic titles.. We summarize our findings below for their cross-cultural patterns and colonial ideologies in pornography.

Our analysis reveals consistent patterns in the ethnic stereotyping and over-sexualization of women. Women are frequently portrayed as objects of male desire, with their value tied to their ethnic and racial background for outgroup members, and their status within domestic frameworks tied to their value as sexual objects for ingroup members pursuing sexual taboos. In both cases, however (ingroup and outgroup), a woman's appearance and her ability to please men is placed at the core of pornographic content with words like \textit{`beauty'}, \textit{`housewife'}, and \textit{`model'} etc. used to describe women. In contrast, men are generally depicted as dominant, masculine figures who engage in physically active, male dominated fields --- \textit{`plumbers'}, \textit{`construction workers'}, etc. In both cases, we note small divergences from the norm where usual masculine and feminine stereotypes are subverted.


The prevalence of stereotyping and exoticizing across different nationalities also suggests that flattening the diverse sexual behaviors, appetites, and needs of women, particularly those from racial and ethnic minorities, is a widespread issue in pornographic discourse, and not simply limited to any single cultural context. It highlights the need for more critical engagement with how women are represented within pornography, and how ethnicity and culture factor into these representations, and what, if any, are the downstream effects on consumers who continue to watch this content.

As noted above, while the hypersexualization of women was observed in all 8 subgroups, the nature and degree of hypersexualization differed based on which ethnic group was under consideration. Our analysis reveals that historical power dynamics, especially sustained through former colonial relationships, continue to shape sexual representation and fantasy creation in pornographic content.
These patterns suggest that pornographic discourse often flattens complex identities into simplistic, stereotypical portrayals that can reinforce harmful biases. While the dataset is extensive, it centers on a single platform, examines only title-level metadata, and focuses on nationality and gender without addressing intersecting identity dimensions such as sexuality, body type, or age. Future research would also benefit from studying the provenance and permeation of pornographic language in the mainstream and gauging its impact on mainstream culture.

\subsection{Limitations and Future Work}

Our analysis focuses solely on nationality as a category, an approach that, though informative, focuses only on a narrow aspect of individual identity in an area (pornography) where a multitude of identities co-exist and interact simultaneously to generate and guide sexual behavior. Future research should examine how nationality intersects with other facets of identity, such as sexual orientation, age, and body type in pornographic content. This would provide a more comprehensive picture of how the influence of various identifiers informs characterization and especially stereotyping within porn.

Additionally, our data comes from a single, albeit major, adult website. A broader analysis across multiple platforms could reveal whether these patterns are consistent across the industry or specific to certain contexts. Longitudinal studies could also shed light on how these representations have evolved over time, revealing what interpretations and characterizations have endured and which ones have been retired. 


Further, as stated above, while our work shows that problematic language exists in porn that reinforces biases present in the real world, the extent to which such language permeates into mainstream discourse remains to be seen. Specifically, a language diffusion analysis can be done to understand pathways of language adaption from porn into mainstream culture. Reddit communities, some of which are centered around the consumption of porn (eg. r/watchitfortheplot), are ideal for such analyses as they represent an opportunity to see how the language of porn has informed mainstream discourse around sex, especially as adult content. 

Finally, the results of this study underscore the need for more critical engagement with pornographic content, not just as a matter of personal consumption, but as a barometer of broader cultural biases and power dynamics. As pornography continues to be a significant part of many people's media consumption, understanding and addressing when and how representation impacts consumers will become increasingly crucial for fostering more equitable and respectful societal attitudes and healthy sexual behaviors.

\bibliography{aaai2026}

\section*{Paper Checklist}
\begin{enumerate}

\item For most authors...
\begin{enumerate}
    \item  Would answering this research question advance science without violating social contracts, such as violating privacy norms, perpetuating unfair profiling, exacerbating the socio-economic divide, or implying disrespect to societies or cultures?
    \answerYes{Yes, and the research focuses on analyzing publicly available data from xHamster and aims to critique the representation of stereotypes in adult content, which aligns with broader ethical concerns regarding media literacy and bias critique. The paper describes how stereotypes are perpetuated rather than reinforcing them.}
  \item Do your main claims in the abstract and introduction accurately reflect the paper's contributions and scope?
    \answerYes{Yes, the abstract and introduction clearly state that the paper addresses the prevalence of ethnic, racial, and gender stereotypes in pornographic discourse.}
   \item Do you clarify how the proposed methodological approach is appropriate for the claims made? 
    \answerYes{Yes, we uses TF-IDF and Ward's method to analyze linguistic patterns, which is appropriate for identifying and quantifying stereotypical language across large datasets. The methods are clearly explained in the methodology section.}
   \item Do you clarify what are possible artifacts in the data used, given population-specific distributions?
    \answerYes{Yes, the paper discusses how the dataset is based solely on title-level metadata and nationality tags, which are limited in capturing the full complexity of ethnic and gender identities. We mention the potential for over-representation of certain nationalities and the limitations of using only xHamster data.}
  \item Did you describe the limitations of your work?
    \answerYes{Yes, the limitations are discussed in the "Limitations and Future Work" section, where the authors mention the narrow focus on nationality, and the use of a single platform.}
  \item Did you discuss any potential negative societal impacts of your work?
    \answerYes{Yes, the paper discusses how its findings could reinforce harmful stereotypes if misinterpreted, and emphasizes the need for critical engagement with pornographic content to avoid legitimizing stereotypes.}
      \item Did you discuss any potential misuse of your work?
    \answerYes{Yes, the potential for reinforcing stereotypes through the paper’s analysis of pornographic language is acknowledged. We have tried our best to avoid amplifying harmful representations.}
    \item Did you describe steps taken to prevent or mitigate potential negative outcomes of the research, such as data and model documentation, data anonymization, responsible release, access control, and the reproducibility of findings?
    \answerYes{Yes, the paper adheres to ethical scraping practices and ensures the data does not include personal or sensitive information. It also provides a link to the notebook used to process the data for transparency.}
  \item Have you read the ethics review guidelines and ensured that your paper conforms to them?
    \answerYes{Yes}
\end{enumerate}

\item Additionally, if your study involves hypotheses testing...
\begin{enumerate}
  \item Did you clearly state the assumptions underlying all theoretical results?
    \answerYes{Yes, the assumptions underlying the analysis are clearly described in the methodology, particularly regarding the use of TF-IDF and the grouping of videos by nationality tags.}
  \item Have you provided justifications for all theoretical results?
    \answerYes{Yes, the results are grounded in theoretical frameworks drawn from existing literature on stereotypes, Orientalism, film theory and the male gaze.}
  \item Did you discuss competing hypotheses or theories that might challenge or complement your theoretical results?
    \answerYes{Yes, we reference previous works that explore similar themes and contrast our findings with other theoretical perspectives.}
  \item Have you considered alternative mechanisms or explanations that might account for the same outcomes observed in your study?
    \answerYes{Yes, alternative explanations, such as cultural differences in media consumption and platform-specific language use, are considered in the paper.}
  \item Did you address potential biases or limitations in your theoretical framework?
    \answerYes{Yes, we acknowledge the potential for biases in our interpretation of the data, especially considering the limitations of analyzing a single platform.}
  \item Have you related your theoretical results to the existing literature in social science?
    \answerYes{Yes, our work builds on existing frameworks in gender studies, and includes discussion on the male gaze, Orientalism, and gender representation in media. We relate our findings to broader discussions of power dynamics and stereotype perpetuation.}
  \item Did you discuss the implications of your theoretical results for policy, practice, or further research in the social science domain?
    \answerYes{Yes, our work calls for more critical engagement with pornography and its societal implications, we argue for further research into media literacy and the effects of pornographic content on public perceptions of gender and race.}
\end{enumerate}

\item Additionally, if you are including theoretical proofs...
\begin{enumerate}
  \item Did you state the full set of assumptions of all theoretical results?
    \answerNA{N/A}
	\item Did you include complete proofs of all theoretical results?
    \answerNA{N/A}
\end{enumerate}

\item Additionally, if you ran machine learning experiments...
\begin{enumerate}
  \item Did you include the code, data, and instructions needed to reproduce the main experimental results (either in the supplemental material or as a URL)?
    \answerYes{Yes, we provide a link to the notebook used for data processing and analysis, which is available at an OSF repository.}
  \item Did you specify all the training details (e.g., data splits, hyperparameters, how they were chosen)?
    \answerNA{We do not train a machine learning model}
     \item Did you report error bars (e.g., with respect to the random seed after running experiments multiple times)?
    \answerNA{N/A}
	\item Did you include the total amount of compute and the type of resources used (e.g., type of GPUs, internal cluster, or cloud provider)?
    \answerNA{N/A}    
     \item Do you justify how the proposed evaluation is sufficient and appropriate to the claims made? 
    \answerNA{N/A}
     \item Do you discuss what is ``the cost`` of misclassification and fault (in)tolerance?
    \answerNA{N/A}
\end{enumerate}

\item Additionally, if you are using existing assets (e.g., code, data, models) or curating/releasing new assets, \textbf{without compromising anonymity}...
\begin{enumerate}
  \item If your work uses existing assets, did you cite the creators?
    \answerYes{Yes, the paper cites relevant datasets and research, including previous studies on pornographic content and language.}
  \item Did you mention the license of the assets?
    \answerNA{N/A}
  \item Did you include any new assets in the supplemental material or as a URL?
    \answerYes{Yes, the dataset and associated analysis notebook are made available via an OSF repository link for reproducibility.}
  \item Did you discuss whether and how consent was obtained from people whose data you're using/curating?
\answerYes{Yes, the data used in the study is publicly accessible and no personally identifiable information is included in the analysis.}  
\item Did you discuss whether the data you are using/curating contains personally identifiable information or offensive content?
    \answerYes{Yes, the data does not contain personally identifiable information and that the analysis focused on public metadata. We do acknowledge that because of the sensitive nature of the topic, some of the content might be offensive or disturbing to some readers.}
\item If you are curating or releasing new datasets, did you discuss how you intend to make your datasets FAIR (see \citet{fair})?
\answerNA{N/A}
\item If you are curating or releasing new datasets, did you create a Datasheet for the Dataset (see \citet{gebru2021datasheets})? 
\answerNA{N/A}

\end{enumerate}

\item Additionally, if you used crowdsourcing or conducted research with human subjects, \textbf{without compromising anonymity}...
\begin{enumerate}
  \item Did you include the full text of instructions given to participants and screenshots?
    \answerNA{NA, this work does not involve human subjects or crowdsourcing.}
  \item Did you describe any potential participant risks, with mentions of Institutional Review Board (IRB) approvals?
    \answerNA{NA, this work does not involve human subjects.}
  \item Did you include the estimated hourly wage paid to participants and the total amount spent on participant compensation?
    \answerNA{NA, this work does not involve human subjects.}
   \item Did you discuss how data is stored, shared, and deidentified?
   \answerNA{NA, this work does not involve human subjects.}
\end{enumerate}

\end{enumerate}

\appendix
\section{Appendix}

\subsection{Data Collection Ethics}
This research involves the systematic scraping of data from the adult website, xHamster, conducted strictly within the platform’s rate limits to minimize any impact on its infrastructure. Our crawls were consistent with typical auditing studies \cite{sandvig2014auditing} and are legally permissible (\cite{hiq_labs_2019}). We adhered to ethical web scraping practices to ensure that our data collection did not interfere with user experience or the platform’s operations.

The analysis of pornographic content, particularly in its portrayal of different nationalities and ethnicities, brings forth significant ethical concerns. By categorizing and analyzing content based on ethnicity, there is a risk of reinforcing harmful stereotypes, precisely the biases we seek to critique. Our goal is to expose how pornographic discourse perpetuates racial, ethnic, and gender biases, but we approach this with caution to avoid inadvertently legitimizing or amplifying these stereotypes.

The data collected from xHamster is publicly accessible and does not contain sensitive personal information such as individual identities, private sexual preferences, or personally identifiable data. Our collection focused solely on title-level and keyword-level metadata related to content portrayal. We took care to handle the data responsibly and followed best practices for privacy protection and ethical research, ensuring that no personal or confidential information was divulged.

\subsection{Data Availability}

The notebook used to process the data can be found at an OSF repository: \url{https://osf.io/kbzx6/?view_only=43b9d8aee7174c02a9cfe80e7f498222}. 
Due to the large size of the dataset, we can make it available through direct communications with the authors.

\end{document}